\documentclass[final]{cvpr}

\usepackage{times}
\usepackage{epsfig}
\usepackage{graphicx}
\usepackage{amsmath}
\usepackage{amssymb}

\usepackage[pagebackref=true,breaklinks=true,colorlinks,bookmarks=false]{hyperref}

\usepackage{overpic}
\usepackage{microtype}
\usepackage{enumitem}
\usepackage{overpic}
\usepackage[below]{placeins}
\usepackage{dsfont}
\usepackage{wrapfig}
\usepackage{booktabs}
\usepackage{balance}

\usepackage{stfloats} 

\usepackage{color}
\definecolor{turquoise}{cmyk}{0.65,0,0.1,0.3}
\definecolor{purple}{rgb}{0.65,0,0.65}
\definecolor{dark_green}{rgb}{0, 0.5, 0}
\definecolor{orange}{rgb}{0.8, 0.6, 0.2}
\definecolor{red}{rgb}{0.8, 0.2, 0.2}
\definecolor{darkred}{rgb}{0.6, 0.1, 0.05}
\definecolor{blueish}{rgb}{0.0, 0.3, .6}
\definecolor{light_gray}{rgb}{0.7, 0.7, .7}
\definecolor{pink}{rgb}{1, 0, 1}
\definecolor{greyblue}{rgb}{0.25, 0.25, 1}
\definecolor{gold}{rgb}{0.7, 0.5, 0}

\renewcommand{\paragraph}[1]{\vspace{.2em}\noindent\textbf{#1}.}

\usepackage{blindtext}

\newcommand{\Figure}[1]{Figure~\ref{fig:#1}}

\newcommand{\Table}[1]{Table~\ref{tab:#1}}
\newcommand{\eq}[1]{\eqref{eq:#1}}
\newcommand{\Eq}[1]{Eq.~\eqref{eq:#1}}

\newcommand{\Section}[1]{Section~\ref{sec:#1}}

\newcommand{\IE}{\mathds{E}}
\newcommand{\IR}{\mathds{R}}

\newcommand{\CIRCLE}[1]{\raisebox{.5pt}{\footnotesize \textcircled{\raisebox{-.6pt}{#1}}}}

\newcommand{\SupplementaryMaterial}{appendix\xspace}

\newcommand{\MethodName}{DeRF\xspace}
\newcommand{\Density}{\sigma}
\newcommand{\Radiance}{\mathbf{c}}
\newcommand{\pos}{\mathbf{x}}
\newcommand{\dir}{\mathbf{d}}
\newcommand{\Ray}{\mathbf{r}}
\newcommand{\PixelColor}{\mathcal{C}}
\newcommand{\Transmittance}{\mathcal{T}}
\newcommand{\NerfParams}{\theta}
\newcommand{\AttentionParams}{\phi}
\newcommand{\Attention}{w}
\newcommand{\PixelAttention}{\mathcal{W}}
\newcommand{\UniformityLoss}{\mathcal{L}_\mathrm{uniform}}
\newcommand{\RadianceLoss}{\mathcal{L}_\mathrm{radiance}}
\newcommand{\Temperature}{\beta}
\newcommand{\Rays}{R}

\definecolor{darkgreen}{rgb}{0,0.6,0}

\newcommand{\bo}{\mathbf{o}}

\begin{document}

\title{DeRF: Decomposed Radiance Fields}

\author{
Daniel Rebain$^1$, \hspace{3pt}
Wei Jiang$^1$, \hspace{3pt}
Soroosh Yazdani$^4$, \hspace{3pt}
Ke Li$^{2, 4}$, \hspace{3pt}
Kwang Moo Yi$^1$, \hspace{3pt}
Andrea Tagliasacchi$^{3, 4}$ \\
\\
$^1$University of British Columbia \hspace{3pt}
$^2$Simon Fraser University \hspace{3pt} \\
$^3$University of Toronto \hspace{3pt} \hspace{3pt}
$^4$Google Research
}

\maketitle

\begin{abstract}
With the advent of Neural Radiance Fields (NeRF), neural networks can now render novel views of a 3D scene with quality that fools the human eye.
Yet, generating these images is very computationally intensive, limiting their applicability in practical scenarios.
In this paper, we propose a technique based on spatial decomposition capable of mitigating this issue.
Our key observation is that there are diminishing returns in employing larger (deeper and/or wider) networks.
Hence, we propose to spatially decompose a scene and dedicate smaller networks for each decomposed part.
When working together, these networks can render the whole scene.
This allows us near-constant inference time regardless of the number of decomposed parts.
Moreover, we show that a Voronoi spatial decomposition is preferable for this purpose, as it is provably compatible with the Painter's Algorithm for efficient and GPU-friendly rendering.
Our experiments show that for real-world scenes, our method provides up to 3$\times$ more efficient inference than NeRF (with the same rendering quality), or an improvement of up to 1.0~dB in PSNR (for the same inference cost).
\end{abstract}

\FloatBarrier
\begin{figure}[t]
\begin{center}
\includegraphics[width=\linewidth]{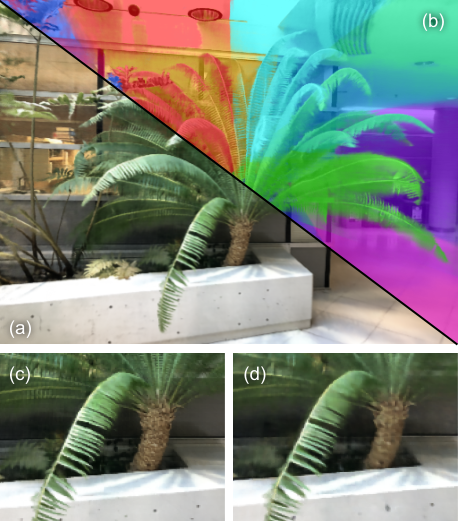}
\end{center}
\caption{
We render a scene~(a) from a decomposed neural representation~(b), consisting of a collection of spatially localized neural networks.
Each of these networks render a \textit{convex} portion of the image, and these are then composited into the output via the Painter's Algorithm.
Depending on the level of decomposition, this can lead to faster rendering, or to renderings that have the same runtime, but contain sharper details.
In~(c), we decompose the scene into 16 parts, which leads to sharper details than (d), with similar runtime.
}
\label{fig:teaser}
\end{figure}
\section{Introduction}
\label{sec:intro}

While high-quality rendering of virtual scenes has long been the sole domain of traditional computer graphics~\cite{pbrt,mitsuba2}, there have recently been promising developments in using neural networks for photo-realistic rendering~\cite{thies2019deferred,neuralvolumes, nritw, nerf}.
These new \emph{neural rendering} methods have the potential to diminish the enormous amount of user intervention that is today needed to digitize the real-world.
We believe the further development of neural scene representations will open the doors to 3D content creation in-the-wild, with the potential to achieve previously unimaginable levels of visual~detail.

Among existing neural rendering methods, those that operate in 3D have lately drawn much interest~\cite{srn, neuralvolumes, nerf}.
Unlike those based on convolutional neural networks~\cite{nsr}, these methods do not operate in image-space, and rather train \textit{volumetric} representations of various types: they define functions that can be queried in space during a volume rendering operation.
This is essential, as volume rendering introduces an \textit{inductive bias} towards rendering phenomena, so that effects like occlusion and parallax are modeled by construction, rather than being emulated by image-space operations.

However, neural volume rendering is far from being a fully developed technology.
The two development axes are \textit{generality}~(removing assumptions about the input, hence allowing their application to more general visual phenomena) and \textit{performance}~(increasing the efficiency of training and/or inference).
In this paper, we focus on performance, and specifically the \textit{inference} performance.
Hence, the natural question then becomes \textit{``why are neural volume rendering models so incredibly slow?''}
Let us consider Neural Radiance Fields (NeRF)~\cite{nerf} as the cardinal example.
These method requires hundreds of MLP invocations per pixel to compute the samples needed by volume rendering.
This results in an extremely compute-intensive inference process needing ${\approx}10^8$ network evaluations and \textit{minutes} of computation to render a one megapixel~image on a modern NVIDIA RTX 2080 Ti accelerator.

\begin{figure}[t]
\begin{center}
\includegraphics[width=\linewidth]{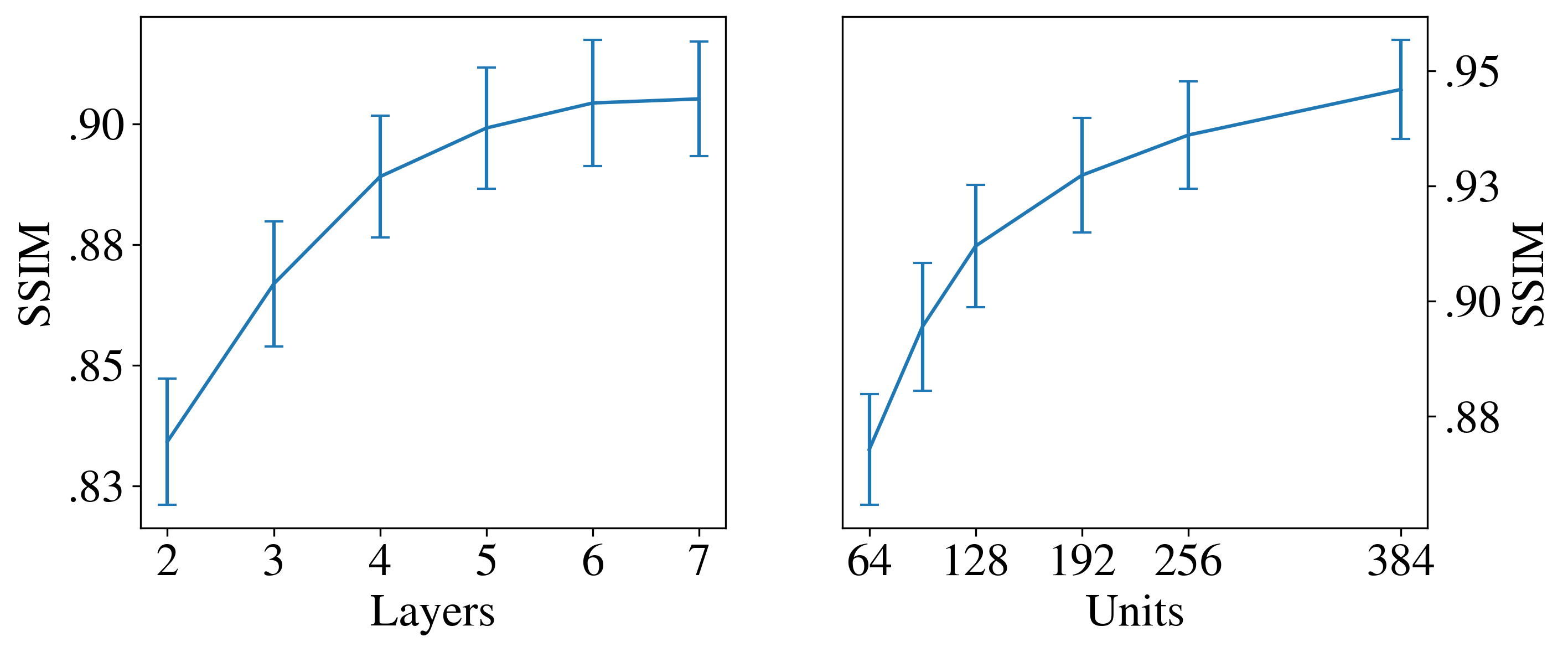}
\end{center}
\vspace{-.15in}
\caption{
\textbf{Diminishing returns} -- We sweep through network architectures varying by depth and width to show how the gains in quality diminish with increased capacity.
The total number of network parameters varies linearly with network depth (left) and quadratically with the number of units in each layer (right).
All networks trained for 300k iterations on the NeRF ``room'' scene.
}
\label{fig:capacityvquality}
\end{figure}
Naturally, to accelerate inference one could trade away model capacity, but doing so na\"{i}vely results in lower rendering quality.
However, as illustrated in \Figure{capacityvquality}, there are \textit{diminishing returns} regarding how the capacity of neural networks~(i.e.~number of layers or number of neurons per layer) affects final rendering quality.
We take advantage of this phenomena and accelerate neural rendering by dividing the scene into \textit{multiple} areas (i.e.~spatial decomposition), and employing a \textit{small(er)} networks in each of these areas.

Due to hardware limitations in the memory architecture of accelerators, not all decompositions are appropriate.
For example, a random decomposition of the volume would result in random sub-network invocations.
Coalescing invocations so that memory access are contiguous is possible, but in our experiments we found the re-ordering operations are not sufficiently fast, and any efficiency gain from using smaller networks was lost.
Hence, our question becomes \textit{``Can we design a spatial decomposition that minimizes the chance of random memory access?''}
We address this question by noting that we can elegantly overcome these limitations if we decompose space with Voronoi Diagrams~\cite[Ch.~7]{voronoi}.
More specifically, the convex cells of the Voronoi diagram can be rendered \textit{independently}, after which the Painter's Algorithm~\cite[Ch.~12]{voronoi} can be used to \textit{composite} the final image.

We formulate our Voronoi decomposition to be differentiable, and train end-to-end to find an optimal cell arrangement.
By doing so we increase the efficiency of the rendering process by up to a factor of three without \textit{any} loss in rendering quality. 
Alternatively, with the same rendering cost, we enhance the rendering quality by a PSNR of up to 1.0dB~(recall that Peak Signal to Noise Ratio is expressed in log-scale).

\paragraph{Contributions}
To summarize, our main contributions are:
\begin{itemize}[leftmargin=*]
\setlength\itemsep{-.3em}
\item We highlight the presence of diminishing returns for network capacity in NeRF, and propose spatial decompositions to address this issue.
\item We demonstrate how a decomposition based on Voronoi Diagrams may be learned to optimally represent a scene.
\item We show how this decomposition allows the whole scene to be rendered by rendering each part independently, and compositing the final image via Painter's Algorithm.
\item In comparison to the NeRF baseline, these modifications result in improvement of rendering quality for the same computational budget, or faster rendering of images given the same visual quality.
\end{itemize}

\section{Related Work}
\label{sec:related}
A large literature exists on neural rendering.
We refer the reader to a recent survey~\cite{star}, and only cover the most relevant techniques in what follows.

\paragraph{Image-space neural rendering}
The simplest form of neural rendering resorts to image-to-image transformations via convolutional neural networks~\cite{nsr}.
This operation can be aided by 3D reasoning~\cite{enr, nvr, nritw, synsin}, producing an intermediate output that is then again fed to a CNN; regular grids~\cite{enr, nvr} or point clouds~\cite{nritw, synsin} have both been used for this purpose.
As these works still rely on CNNs to post-process the output, they have difficulty modeling view-dependent effects, often resulting in visible artefacts.

\begin{figure*}
\begin{center}
\includegraphics[width=.54\textwidth]{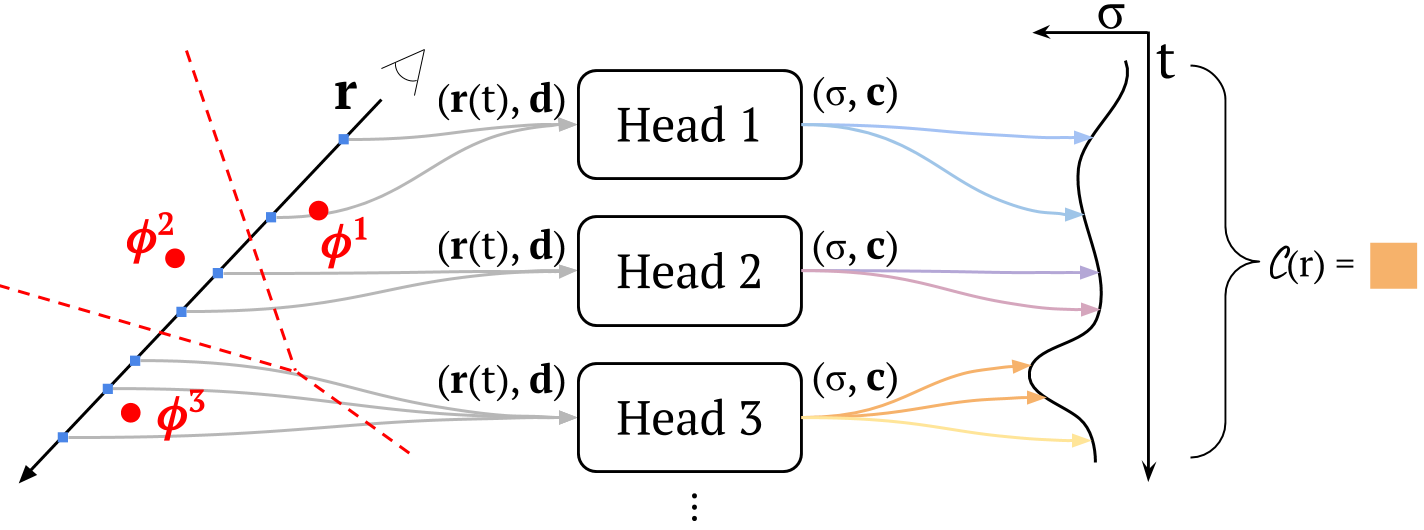}
\includegraphics[width=.44\textwidth]{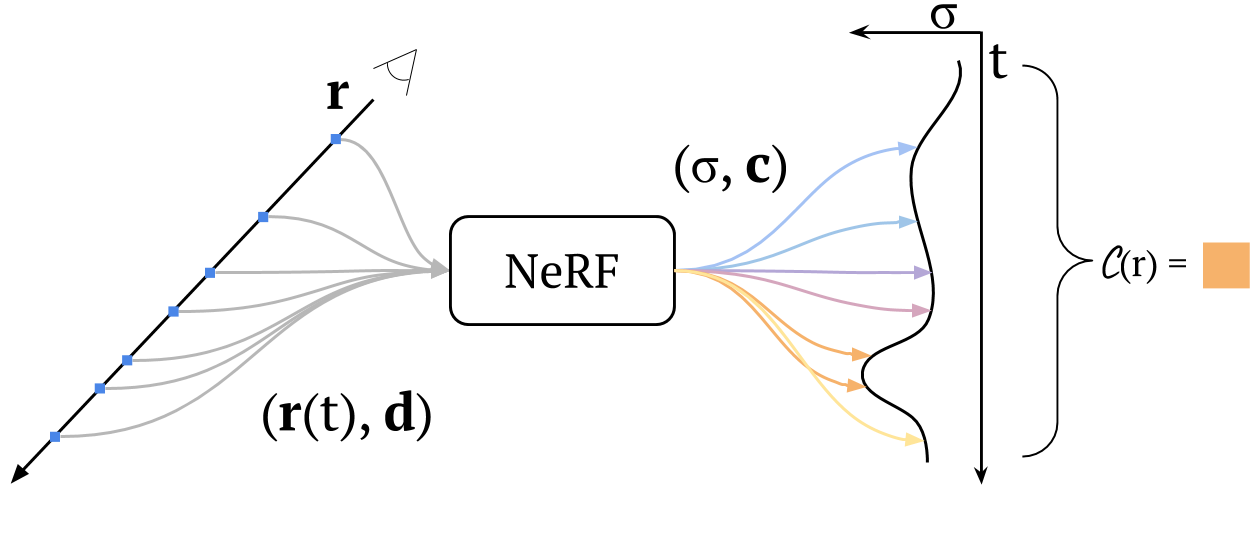}
\end{center}
\caption{
\textbf{Framework -- }
The \MethodName~architecture (left) consists of a set of independent NeRF (right) networks which are each responsible for the region of space within a Voronoi cell defined by the decomposition parameters $\AttentionParams$. The final color value for a ray is computed by applying the volume rendering equation to each segment of radiance $\Radiance$ and density $\Density$, and alpha compositing together the resulting colors.
}
\label{fig:framework}
\end{figure*}
\begin{figure*}[t]
\begin{center}
\includegraphics[width=\linewidth]{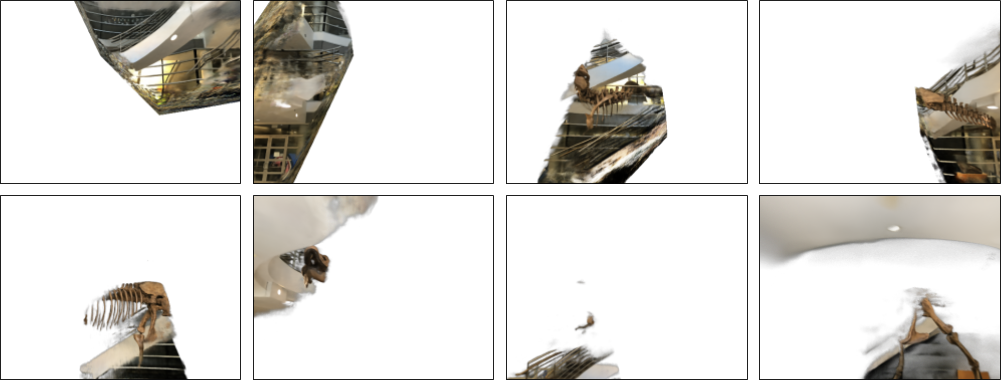}
\end{center}
\vspace{-1em}
\caption{
\textbf{Decomposed radiance fields} --
We visualize each of the rendering heads individually.
Note that as each head is rendered \textit{only} the weights of \textit{one} neural network head needs to be loaded, hence resulting in optimal cache coherency while accessing GPU memory.
}
\label{fig:layers}
\end{figure*}
\paragraph{Neural volumetric rendering}
Recently, researchers succeeded in integrating 3D inductive bias within a network in a completely end-to-end fashion, hence removing the need CNN post-processing.
Instead, they rely on tracing rays through a volume to render an image~\cite{deepvoxels, srn}.
While these results pioneered the field, more compelling results were achieved via the use of fixed-function volume rendering~\cite{neuralvolumes, nerf}.
In particular, and thanks to the use of positional encoding, NeRF~\cite{nerf} is able to render novel views of a scene from a neural representation with photo-realistic quality.
Extensions of NeRF to dynamic lighting and appearance exist~\cite{nitw}, as well as early attempts at decomposing the complexity of the scene into near/far components~\cite{nerfplusplus}.
With an objective similar to ours, in Neural Sparse Voxel Fields~\cite{nsvf}, the authors realize a 10$\times$ speed-up by discretizing the scene and avoiding computation in empty areas; note this solution is \textit{complementary} to ours. 
It focuses on the \textit{sampling} part of the NeRF pipeline, and therefore can be used in conjunction with what we propose.
\begin{figure*}[t]
\begin{center}
\includegraphics[width=\linewidth]{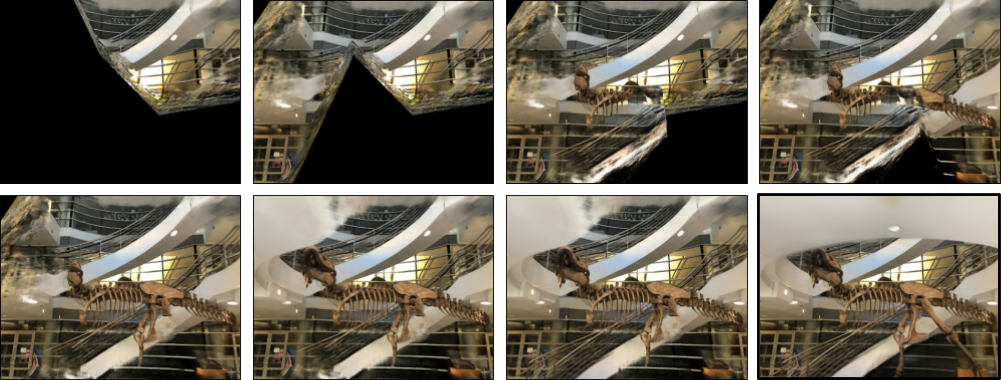}
\end{center}
\vspace{-1em}
\caption{
\textbf{Compositing with the Painter's algorithm} --
Visualization of the intermediate steps of the compositing process.
}
\label{fig:painters}
\end{figure*}
\section{Method}
\label{sec:method}
We review the fundamentals of NeRF in~\Section{background}, describe our decomposition-based solution in~\Section{decomposition}, its practical realization with Voronoi Diagrams and the Painter's Algorithm in~\Section{voronoi}.
We conclude by detailing our training methodology in~\Section{implementation}.

\subsection{Neural radiance fields (NeRF)}
\label{sec:background}
To represent a scene, we follow the volume rendering framework of NeRF~\cite{nerf}; see \Figure{framework}~(right).
Given a camera ray $\Ray(t) = \bo + t\dir$ corresponding to a \textit{single} pixel, we integrate the contributions of a 5D~(3D space plus 2D for direction) radiance field $\Radiance(\pos, \dir)$ and spatial density~$\Density(\pos)$ along the ray:
\begin{equation}
\PixelColor(\Ray) = \int_{t_n}^{t_f} \Transmittance(t) \: \Density(\Ray(t)) \: \Radiance(\Ray(t), \dir) \, dt
\end{equation}
to obtain a the pixel color $\PixelColor(\Ray)$. Here, $t_n$ and $t_f$ are the near/far rendering bounds, and transmittance $\Transmittance(t)$ represents the amount of the radiance from position $t$ that will make it to the eye, and is a function of density:
\begin{equation}
\Transmittance(t) = \mathrm{exp}\left(-\int_{t_n}^{t} \Density(\Ray(s)) \, ds \right)    
\end{equation}
The neural fields $\Density(\pos)$ and $\Radiance(\pos, \dir)$ are trained to minimize the difference between the rendered and observed pixel values $\PixelColor_\mathrm{gt}$ over the set of all rays $\Rays$ from the training images:
\begin{equation}
\RadianceLoss = \IE_{\Ray\sim\Rays}\left[\left\|\PixelColor(\Ray) - \PixelColor_\mathrm{gt}(\Ray)\right\|_2^2\right]
\;.
\end{equation}
Note how in neural radiance fields, a \textit{single} neural network is used to estimate $\Density(\pos)$ and $\Radiance(\pos, \dir)$ for the \textit{entire} scene.
However, as discussed in the introduction, it is advisable to use multiple smaller capacity neural networks (heads) to compensate for diminishing returns in rendering accuracy.

\subsection{Decomposed radiance fields (DeRFs)}
\label{sec:decomposition}
We propose to model the radiance field functions $\Density(\pos)$ and $\Radiance(\pos, \dir)$ as a weighted sum of $N$ separate functions, each represented by a neural network~(head); see \Figure{framework}~(left).
Specifically, the NeRF model defines two directly learned functions: $\Density_\NerfParams(\pos)$ and $\Radiance_\NerfParams(\pos, \dir)$, each defined for values of $\pos$ over the full space of $\mathbb{R}^3$, and modeled with a neural network with weights $\NerfParams$.
Conversely, in \MethodName{} we~write:
\begin{align}
\Density(\pos) &= \sum\nolimits_{n=1}^N \Attention_{\AttentionParams}^n(\pos) \Density_{\NerfParams_n}(\pos)\\
\Radiance(\pos, \dir) &= \sum\nolimits_{n=1}^N  \Attention_{\AttentionParams}^n(\pos) \Radiance_{\NerfParams_n}(\pos, \dir)
\label{eq:decomposed}
\end{align}
where $n$ denotes the head index, and $\Attention_{\AttentionParams}(\pos) {:} \IR^3 {\mapsto} \IR^N$ represents our decomposition via a learned function~(with parameters $\AttentionParams$) that is coordinatewise positive and satisfies the property $\|\Attention_\AttentionParams(\pos)\|_1 {=} 1$.

\paragraph{Efficient scene decomposition}
Note how in \Eq{decomposed}, whenever $\Attention_{\AttentionParams}^n(\pos){=}0$ there is no need for $\Density_{\NerfParams_n}(\pos)$ and $\Radiance_{\NerfParams_n}(\pos, \dir)$ to be evaluated, as their contributions would be zero.
Hence, we train our decomposition $\Attention_{\AttentionParams}$ so that only \textit{one} of the N elements in $\{\Attention_\AttentionParams^n(\pos)\}$ is non-zero at any position in space~(i.e.~we have a spatial partition).
Because of this property, for each $\pos$, only \textit{one} head needs to be evaluated, accelerating the inference process.

\paragraph{Balanced scene decomposition}
As all of our heads have similar representation power, it is advisable to decompose the scene in a way that all regions represent a similar amount of information~(i.e.~visual complexity).
Toward this objective, we first introduce $\PixelAttention_\AttentionParams(\Ray) \in \IR^N$ to measure how much the $N$ heads contributes to a given ray:
\begin{equation}
  \PixelAttention_\AttentionParams(\Ray) = \int_{t_n}^{t_f} \Transmittance(t) \: \Density(\Ray(t)) \:  \Attention_\AttentionParams(\Ray(t)) \, dt  
  \;.
  \label{eq:contrib}
\end{equation}
and employ a loss function that enforces the contributions to be uniformly spread across the various heads:
\begin{equation}
\UniformityLoss = \left\| \IE_{\Ray\sim\Rays}\left[ \PixelAttention_\AttentionParams(\Ray) \right]\right\|_2^2
\;.
\end{equation}
 
Minimizing this loss results in a decomposition which utilizes all heads equally.
To see this, let 
$\textbf{W}_\AttentionParams = \IE_{\Ray\sim\Rays} [\PixelAttention_\AttentionParams(\Ray)]$,
and let $\mathbf{1} \in \mathbb{R}^N$ be the vector with all $1$'s.
Recall that since $\|\Attention_\AttentionParams(\pos)\|_1 {=} 1$ and is coordinatewise positive, we get that 
$\mathbf{1} \cdot \Attention_\AttentionParams(\pos) =1 $.
Therefore, $\mathbf{1} \cdot \PixelAttention_\AttentionParams(\Ray)$ will be independent of $\AttentionParams$, and hence $\mathbf{1} \cdot \textbf{W}_\AttentionParams$ will be a constant.
Finally, by the Cauchy-Schwartz inequality, we get that $\| \mathbf{1} \|_2^2 \|\textbf{W}_\AttentionParams\|_2^2$ is minimized when $\textbf{W}_\AttentionParams$ is
parallel to $\mathbf{1}$, which means that all heads contribute equally.

\subsection{Voronoi learnable decompositions}
\label{sec:voronoi}
We seek a decomposition satisfying these requirements:
\vspace{-.25em}
\begin{enumerate}
\setlength\itemsep{-.3em}
\item it must be differentiable, so that the decomposition can be fine-tuned to a particular scene
\item the decomposition must be a spatial \textit{partition} to unlock efficient evaluation, our core objective
\item it must be possible to evaluate the partition in an accelerator-friendly fashion 
\end{enumerate}
\vspace{-.25em}
Towards this objective, we select a Voronoi Diagram as the most suitable representation for our decomposition.
We employ the solution proposed in \cite{voronoinet}, which defines, based on a set $\AttentionParams \in \mathbb{R}^{N \times 3}$ of $N$ Voronoi sites, a differentiable~(i.e.~soft) Voronoi Diagram as:
\begin{equation}
\Attention_\AttentionParams^n(\pos) = \frac{e^{-\Temperature ||\pos - \AttentionParams^n||_2}}{\sum_{j=1}^N e^{-\Temperature ||\pos - \AttentionParams^j||_2}}
\label{eq:voronoi}
\end{equation}
where $\Temperature \in \mathbb{R}^+$ is a temperature parameter controlling the softness of the Voronoi approximation.
This decomposition is:
\CIRCLE{1} differentiable w.r.t. its parameters $\AttentionParams$, and has smooth gradients thanks to the soft-min op in~\eq{voronoi}; 
\CIRCLE{2} a spatial partition for $\Temperature\rightarrow\infty$, and thanks to the defining characteristics of the Voronoi diagram;
\CIRCLE{3} compatible with the classical ``Painter's Algorithm'', enabling efficient compositing of the 
rendering heads to generate the final image.
An example of the trained decomposition is visualized in~\Figure{layers}.

\paragraph{Painter's Algorithm}
\label{sec:painters}
This algorithm is one of the most elementary rendering techniques; see~\cite[Ch.~12]{voronoi}.
\begin{wrapfigure}{r}{0.4\linewidth}
\vspace{-1em}
    \begin{center}
    \includegraphics[width=\linewidth]{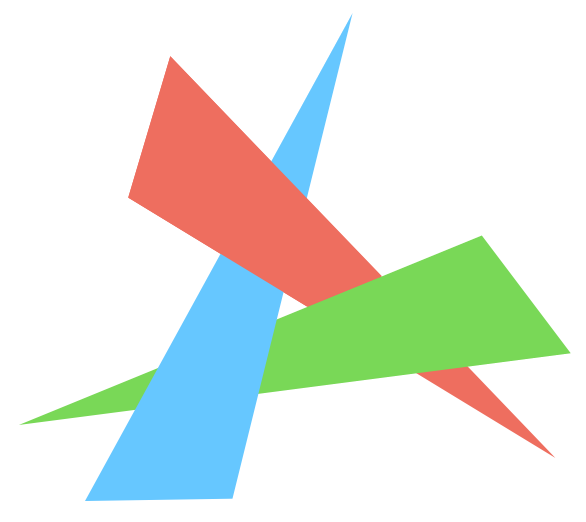}
    \end{center}
    \caption{A simple scene made of convex elements that the Painter's Algorithm cannot render.}
    \label{fig:unsortable}
\end{wrapfigure}
The idea is to render objects independently, and draw them \textit{on top} of each other in order, from back to front, to the output buffer~(i.e.~image).
Such an ordering ensures that closer objects occlude further ones.
This algorithm has only found niche applications (e.g.~rendering transparencies in real-time graphics), but it encounters failure cases when the front to back ordering of objects in the scene is non decidable; see a simple example in the inset figure, where three convex elements form a cycle in the front/behind relation graph.
\par In our solution, the scene can be rendered part-by-part, one Voronoi cell at a time, without causing memory cache incoherences, leading to better GPU throughput.
We can then composite the cell images back-to-front, via the Painter's Algorithm, to generate the final image; see~\Figure{painters}.
However, we need to verify that 
Voronoi decompositions are compatible with the Painter's Algorithm.

\paragraph{On the correctness of the Voronoi Painter's} 
Taking an approach similar to \cite{edelsbrunner1990acyclicity}, we prove that our Voronoi decomposition is compatible with the Painter's Algorithm.
To do so, we will show that for any Voronoi decomposition and a camera
located at $Q$, there is a partial ordering
of the Voronoi cells so that if $V$ shows up before $W$ in our ordering, then $W$ does not occlude any part of $V$ (i.e.
no ray starting from $Q$ to any
point in $V$ will intersect $W$). 
To this end, let $\mathcal{P} \subset \mathbb{R}^n$ be a set of points, and for each $P \in \mathcal{P}$ let
$V_P$ be the Voronoi cell attached to $P$.
For any $Q \in \mathbb{R}^n$ define $<_Q$ on the Voronoi cells of 
\begin{equation}
    V_{P'} <_Q V_P \, \textrm{ if and only if }\, d(P',Q) < d(P,Q),
\end{equation}
where $d$ defines the distance between points.
This clearly defines a partial ordering on $\mathcal{P}$. We now show that this partial ordering is the
desired partial ordering for the Painter's Algorithm. Let $(x, x') \in V_P \times V_{P'}$ and let
$x' = \lambda x + (1-\lambda) Q$ for $\lambda \in (0, 1)$ (i.e. $x'$ is on the line segment $(x,Q)$) and
hence parts of $V_{P'}$ is covering $V_P$.
We now need to show that $V_{P'} <_Q V_P$, or equivalently $d(P', Q) < d(P,Q)$. 
Let 
\begin{equation}
H = \left\{ z \, | \, d(z,P) < d(z,P')\right \}
\end{equation}
be the halfspace of points closer to $P$.
Note that $d(x,P) < d(x,P')$ (since $x \in V_P$) and $d(x',P) > d(x', P')$,
and hence $x \in H$ and $x' \not \in H$. If $Q \in H$
then the line segment $(x,Q)$ will intersect the boundary of $H$ twice, once between $(x,x')$ and once between
$(x', Q)$, which is not possible. Therefore $Q \not \in H$, which implies $d(P',Q) < d(P,Q)$ as desired.

\subsection{Training details}
\label{sec:implementation}
When training our model, we find that successful training is only achieved when the decomposition function $\Attention_\AttentionParams$ is trained \emph{before} the network heads for density $\Density_{\NerfParams_n}$ and radiance $\Radiance_{\NerfParams_n}$.
This allows the training of the primary model to proceed without interference from shifting boundaries of the decomposition.
However, as shown in \Eq{contrib}, the training of $\Attention_\AttentionParams$ requires a density model $\Density$.
To resolve this problem, we first train coarse networks $\Density_\mathrm{coarse}$ and $\Radiance_\mathrm{coarse}$ that apply to the entire scene, before the networks heads $\Density_{\NerfParams_n}$ and $\Radiance_{\NerfParams_n}$ are trained; this pre-training stage lasts $\approx$100k iterations in our experiments.
During the pre-training stage, to stabilize training even further, we optimize $\NerfParams_\mathrm{coarse}$ and $\AttentionParams$ \textit{separately}, respectively minimizing the reconstruction loss $\RadianceLoss$ and uniformity loss $\UniformityLoss$.
We found through experiment that allowing $\UniformityLoss$ to affect the optimization of $\NerfParams_\mathrm{coarse}$ inhibited the ability of the density network to properly learn the scene structure, resulting in unstable training.
Once $\Attention_\AttentionParams$ is pre-trained, we keep $\AttentionParams$ fixed and train the per-decomposition networks $\Density_{\NerfParams_n}$ and $\Radiance_{\NerfParams_n}$ with the $\RadianceLoss$, as the Voronoi sites are fixed, and $\UniformityLoss$ is no longer is necessary.

\paragraph{Controlling the temperature parameter}
The soft-Voronoi diagram formulation in \eq{voronoi} leads to differentiability w.r.t the Voronoi sites (as otherwise gradients of the weight function would be zero), but efficient scene decomposition~(\Section{decomposition}) requires spatial \textit{partitions}.
Hence, we define a scheduling for $\Temperature$ over the training process.
We start at sufficiently low value so that $\Attention^i(\pos) \approx \Attention^j(\pos)$ for all $i$ and $j$.
As the training progresses, we exponentially increase $\Temperature$ until it reaches a sufficiently high value ($10e9$ in our experiments), so that the decomposition is indistinguishable from a (hard) Voronoi diagram; i.e. either $\Attention^i(\pos) \approx 1$ or $\Attention^i(\pos) \approx 0$~for~all~$i$.

\paragraph{Parameter magnitude scaling}
When using an optimizer such as ADAM~\cite{adam}, the training process is not necessarily invariant with respect to the magnitude of individual variables.
As a result, when using a common optimizer state for all trainable variables, we found that scaling the scene coordinate system such that $\AttentionParams \in [-1, 1]^{N \times 3}$ was necessary to prevent the training of the decomposition from oscillating excessively or progressing too slowly.

\begin{figure*}
\begin{center}
\includegraphics[width=\linewidth]{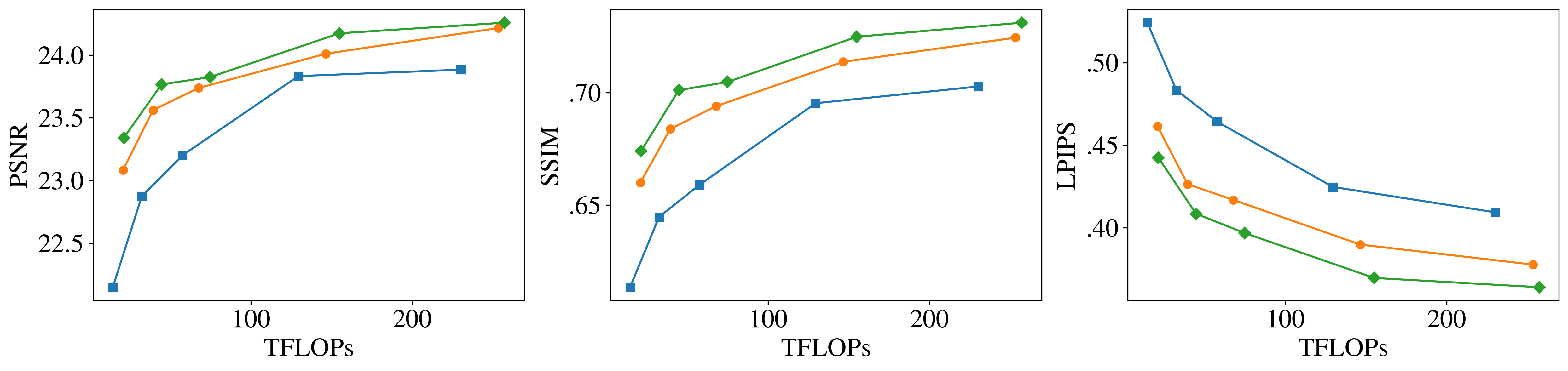}
\end{center}
\vspace{-.25in}
\begin{center}
\resizebox{.8\linewidth}{!}{
\begin{tabular}{c|ccc|ccc|ccc}
\toprule
\multicolumn{1}{l|}{Units} & \multicolumn{3}{c|}{1 Head} & \multicolumn{3}{c|}{4 Heads} & \multicolumn{3}{c}{8 Heads} \\
\multicolumn{1}{l|}{}      & PSNR $\uparrow$ & SSIM$\uparrow$ & LPIPS $\downarrow$ & PSNR$\uparrow$ & SSIM$\uparrow$ & LPIPS $\downarrow$ & PSNR$\uparrow$ & SSIM$\uparrow$ & LPIPS $\downarrow$ \\
\midrule
64  & 22.15 & 0.61 & 0.52 & 23.08 & 0.66 & 0.46 & 23.34 & 0.67 & 0.44 \\
96  & 22.87 & 0.64 & 0.48 & 23.56 & 0.68 & 0.43 & 23.77 & 0.70 & 0.41 \\
128  & 23.20 & 0.66 & 0.46 & 23.74 & 0.69 & 0.42 & 23.82 & 0.70 & 0.40 \\
192  & 23.83 & 0.70 & 0.42 & 24.01 & 0.71 & 0.39 & 24.17 & 0.72 & 0.37 \\
256  & 23.88 & 0.70 & 0.41 & 24.22 & 0.72 & 0.38 & \bf{24.26} & \bf{0.73} & \bf{0.36} \\
\bottomrule
\end{tabular}
} 
\end{center}
\caption{
\textbf{Quality vs. efficiency} --
Reconstruction quality versus run-time inference cost for the ``fern'' scene as the network capacity~(number of hidden units) is changed; the table report the data used to draw the diagrams.
To make the computational requirements tractable, we lower sample counts (128 per ray) and batch sizes (512) are used than for results reported in~\cite{nerf}, and thus are not directly comparable.
For quantitative results for other scenes, please refer to the \SupplementaryMaterial.
}
\label{fig:qualityvefficiency}
\end{figure*}

\section{Results}
\label{sec:experiments}

We now provide our empirical results.
We first detail the experimental setup in \Section{exp_setup}.
We then present how the theory of our method translates to practice, and how our method performs under various computation loads in \Section{exp_main}.
We then discuss other potential decomposition strategies in \Section{exp_decomp}.

\subsection{Experimental setup}
\label{sec:exp_setup}
To validate the efficacy of our method, we use the ``Real Forward-Facing'' dataset from NeRF~\cite{nerf}.
The dataset is composed of eight scenes, which includes 5 scenes originally from \cite{llff}, each with a collection of high-resolution images, and corresponding camera intrinsics and extrinsics.
As we are interested in the relationship between rendering quality and the computational budget in a practical scenario, we focus on real images.
For results on the NeRF~\cite{nerf} synthetic and DeepVoxels~\cite{deepvoxels} synthetic datasets, see \SupplementaryMaterial.

\paragraph{Implementation}
We implement our method in in TensorFlow~2~\cite{tensorflow}.
Due to the large number of evaluation jobs, we train with some quality settings reduced: we use a batch size of 512 and 128~samples/ray. We train each model for 300k iterations~(excluding the decomposition pre-training).
Other settings are the same as reported in NeRF~\cite{nerf}

\subsection{Efficiency of DeRF}
\label{sec:exp_main}

\begin{table}[t]
\footnotesize
\begin{center}
\resizebox{\linewidth}{!}{
\begin{tabular}{c|cc|cc|cc}
\toprule
\multicolumn{1}{l|}{Units} & \multicolumn{2}{c|}{1 Head} & \multicolumn{2}{c|}{4 Heads} & \multicolumn{2}{c}{8 Heads} \\ 
\multicolumn{1}{l|}{} & TFLOPs $\downarrow$ & Time $\downarrow$ & TFLOPs $\downarrow$ & Time $\downarrow$ & TFLOPs $\downarrow$ & Time $\downarrow$ \\
\midrule
64  & \bf{14.5} & \bf{14.4} & 19.6 & 20.7 & 21.1 & 21.2 \\
96  & 26.0 & 32.4 & 29.2 & 39.3 & 37.7 & 44.6 \\
128  & 34.4 & 57.5 & 46.0 & 67.6 & 55.9 & 74.7 \\
192  & 83.3 & 129.4 & 87.2 & 146.5 & 111.0 & 155.0 \\
256  & 115.7 & 230.1 & 160.9 & 253.5 & 186.6 & 257.4 \\
\bottomrule
\end{tabular}
} 
\end{center}
\caption{
Quantitative comparison of average TeraFLOPs and seconds needed to generate a frame for the 
``fern'' scene on an NVIDIA v100 GPU; for both metrics smaller values are better.
}
\label{tab:efficiency}
\end{table}
\paragraph{Theory vs. practice  -- \Table{efficiency}}
In theory the decomposition should not cause any significant increase in computation -- the only increase would be to decompose and then merge rendering results.
We validate that this is in fact the case by evaluating the number of rendered frames-per-TeraFLOP~(theoretical performance) as well as the number of frames-per-second~(practical performance). 
As shown in \Table{efficiency}, in terms of FLOPs, the difference between using no-decomposition and 8-decompositions is 62.8\% in maximum, and 49.8\% in average, whereas there is a quadratic increase in computation should more units be used.

These theoretical trends are mirrored into those of actual runtime.
While there can be an increase in computation time compared to the operation count, the maximum increase when going from no-decomposition to 8-decomposition is 47.5\%, and the average is 29.3\%.
Again, this is much more efficient than increasing the number of neurons.
This highlights the efficacy of our decomposition strategy -- Voronoi -- allowing theory to be applicable in practice.
Note that in \Section{exp_decomp}, we show that this is not necessarily the case for other na\"ive strategies.

\begin{figure*}
\begin{center}
\includegraphics[width=\linewidth]{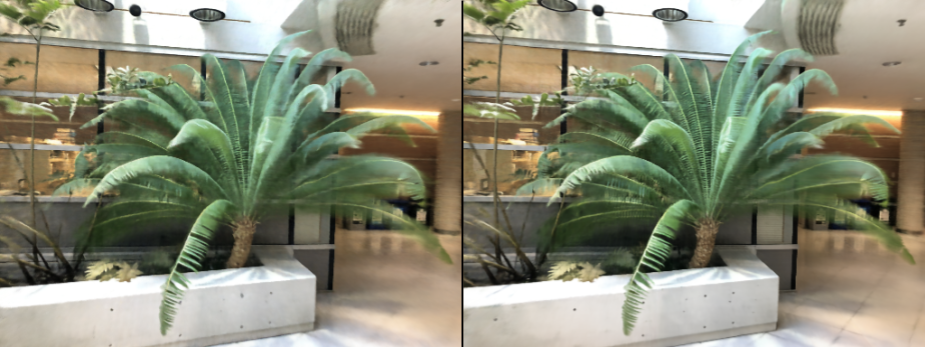}
\end{center}
\vspace{-1em}
\caption{
\textbf{Qualitative comparison} -- 
A direct comparison of the results of a standard NeRF model with 256 hidden units (left), and a 16-head \MethodName~model with 128 units (right).
The \MethodName~model has better \textit{performance} and \textit{quality} than the baseline while using networks which each have one quarter the number of parameters.
In other words, despite our \MethodName{} having $16\times \tfrac{1}{4}=4\times$ more parameters than the baseline NeRF, it executes 1.7$\times$ faster.
See~\Figure{qualityvefficiency} for metrics on various model combinations for this scene.
}
\label{fig:rgbcomparison}
\end{figure*}
\paragraph{Quality vs. efficiency -- \Figure{qualityvefficiency}}
We further evaluate how the quality of rendering changes with respect to the number of decompositions, and the number of neurons used.
To quantify the rendering quality, we rely on three metrics:
\vspace{-.5em}
\begin{itemize}[leftmargin=*]
\setlength\itemsep{-.3em}
\item Peak Signal to Noise Ratio (PSNR): A classic metric to measure the corruption of a signal.
\item Structural Similarity Index Measure (SSIM) \cite{ssim}: A perceptual image quality assessment based on the degradation of structural information.
\item Learned Perceptual Image Patch Similarity (LPIPS) \cite{lpips}: A perceptual metric based on the deep features of a trained network that is more consistent with human judgement.
\end{itemize}
\vspace{-.5em}
We summarize the results for a representative scene in \Figure{qualityvefficiency}.
As shown, given the same render cost, more fine-grained decompositions improve rendering quality across all metrics.
Regardless of the computation, using more decomposition leads to better rendering quality.

\begin{figure*}
\begin{center}
\includegraphics[width=\linewidth]{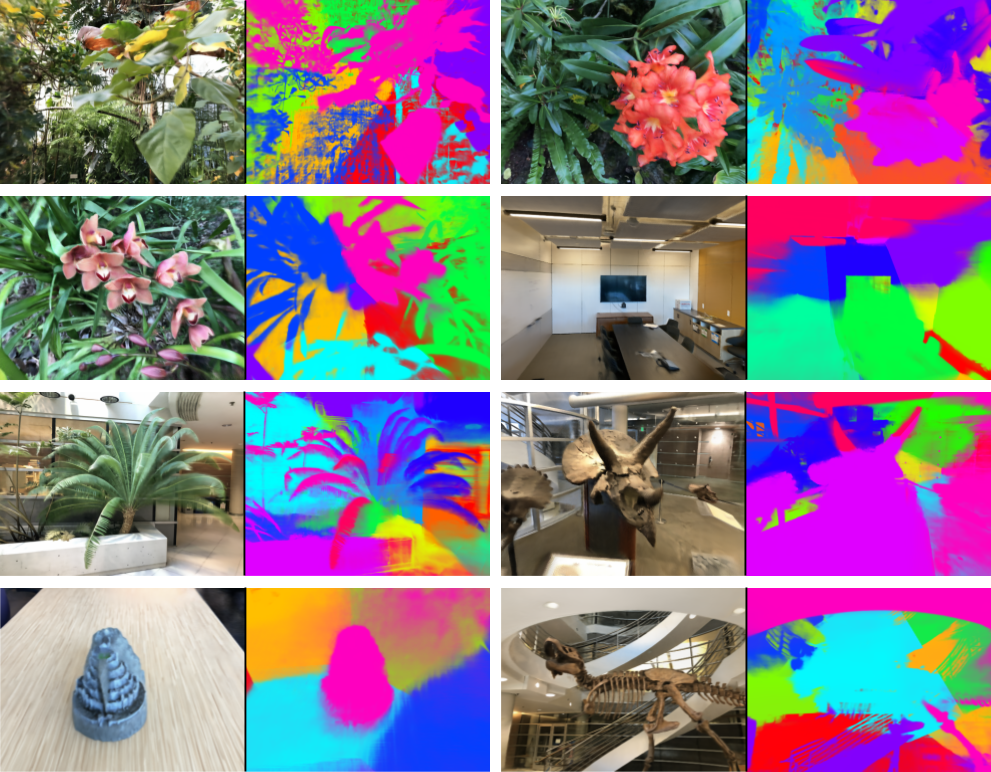}
\end{center}
\vspace{-1em}
\caption{
\textbf{Qualitative results gallery} -- 
A sampling of \MethodName renders, alongside visualizations of decompositions.
}
\label{fig:gallery}
\vspace{-0.8em}
\end{figure*}

\paragraph{Qualitative results -- \Figure{rgbcomparison} and \Figure{gallery}}
We further show a qualitative comparison between a standard NeRF model and DeRF in \Figure{rgbcomparison}, where we show that our method outperforms NeRF in terms of both quality and efficiency.
More qualitative results are also available in \Figure{gallery} and the video supplementary.

\subsection{Alternative decomposition methods}
\label{sec:exp_decomp}

We further empirically demonstrate that na\"ive decomposition strategies are insufficient, and hence the importance of using our Voronoi decomposition.

\begin{table}[t]
\begin{center}
\includegraphics[width=0.8\linewidth]{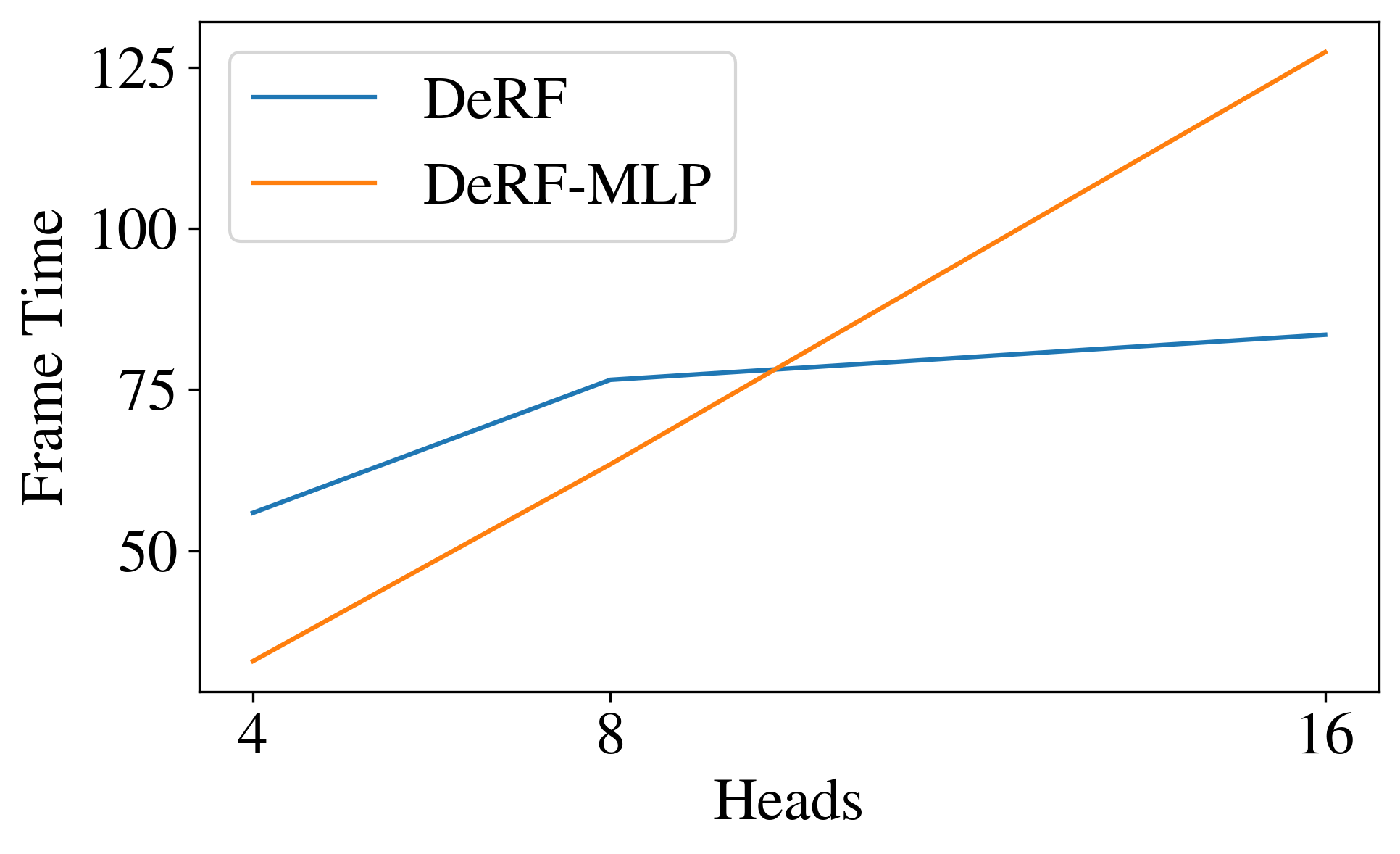}
\end{center}
\setlength{\tabcolsep}{5pt}
\begin{center}
\vspace{-.1in}
\begin{tabular}{l|rrr}
\toprule
 & 4 Heads & 8 Heads & 16 Heads \\
\midrule
\MethodName & 55.88s & 76.51s & \textbf{83.50s} \\
\MethodName-MLP & \textbf{32.89s} & \textbf{63.40s} & 127.32s \\
\bottomrule
\end{tabular}
\end{center}
\caption{
\textbf{Decomposition baselines --}
The frame render time for the ``room'' scene (in seconds) for a Voronoi and MLP-decomposition.
Note that as the number of heads increases, the Voronoi decomposition provides a substantial benefit to efficiency.
Note models beyond 16 heads are infeasible to train due to memory requirements~(for reasonable network widths).
}
\label{tab:mlpablation}
\end{table}
\paragraph{Decompositions with MLPs -- \Table{mlpablation}}
An obvious first thought into decomposing scenes would be to leave the decomposition to a neural network, and ask it to find the optimal decomposition through training.
To compare against this baseline, we implement a decomposition network with
an MLP with a softmax activation at the end to provide values of $\Attention_{\AttentionParams}^n(\pos)$.
We show the actual rendering time compared to ours in \Table{mlpablation}. 
While in theory this method should require a similar number of operations to ours, due to the random memory access during the integration process, they can not be accelerated as efficiently. Consequently, their runtime cost tends to grow much faster than \MethodName models as we increase the number of heads.

\paragraph{Decompositions with regular grids}
A trivial spatial decomposition could be achieved by using a regular grid of network regions.
While this would eliminate the requirement to train the decomposition, in practice it would require many more regions in total to achieve the same level of accuracy (due to the non-homogeneous structure of real scenes).
Due to the curse of dimensionality, this will also result in a significant amount of incoherence in the memory access pattern for network weights, resulting in lower computational performance; see the~\SupplementaryMaterial.

\section{Conclusions}
\label{sec:conclusion}
We have presented DeRF~–~\textit{Decomposed Radiance Fields}~–~a method to increase the inference efficiency of neural rendering via spatial decomposition.
By decomposing the scene into multiple cells, we circumvent the problem of diminishing returns in neural rendering: increasing the network capacity \textit{does not} directly translate to better rendering quality.
To decompose the scene, we rely on Voronoi decompositions, which we prove to be compatible with the Painter's algorithm, making our inference pipeline GPU-friendly.
As a result, our method not only renders much faster, but can also deliver higher quality images.

\paragraph{Limitations and future work}
There are diminishing returns with respect to the number of decomposition heads, not just network capacity; see \Figure{qualityvefficiency}.
Yet, one is left to wonder whether the saturation in rendering quality could be compensated by significantly faster rendering as we increase the number of heads in the hundreds or thousands.
In this respect, while in this paper we assumed all heads to have the same neural capacity, it would be interesting to investigate heterogeneous DeRFs, as, \eg, a 0-capacity head is perfect for representing an empty space.
On a different note, the implementation of highly efficient \textit{scatter}/\textit{gather} operations could lead to an efficient version of the simple MLP solution in~\Section{exp_decomp}, and accelerate the training of DeRFs, which currently train slower than models without decomposition.

\section*{Acknowledgements}
This work was supported by the Natural Sciences and Engineering Research Council of Canada (NSERC) Discovery Grant, NSERC Collaborative Research and Development Grant, Google, Compute Canada, and Advanced Research Computing at the University of British Columbia.

We thank Ricardo Martin Brualla for his comments.

{
    \small
    \bibliographystyle{ieee_fullname}
    \bibliography{macros,main}
}
\clearpage
\clearpage
\twocolumn[
\centering
\Large
\textbf{Appendix}
\vspace{1.0em}
]
\appendix

\balance

\section{Voronoi vs na\"ive decomposition -- \Table{gridablation}}

To test that our learned decomposition actually contributes to an improvement in accuracy, we test against a version of DeRF with a na\"ive decomposition -- where cells are fixed into a grid. 
We find that the learned decomposition results in improvement in all metrics.

\section{Additional experiments on real data}

\begin{figure*}[b]
\vspace{-1em}
\footnotesize
\begin{center}
\includegraphics[width=\linewidth]{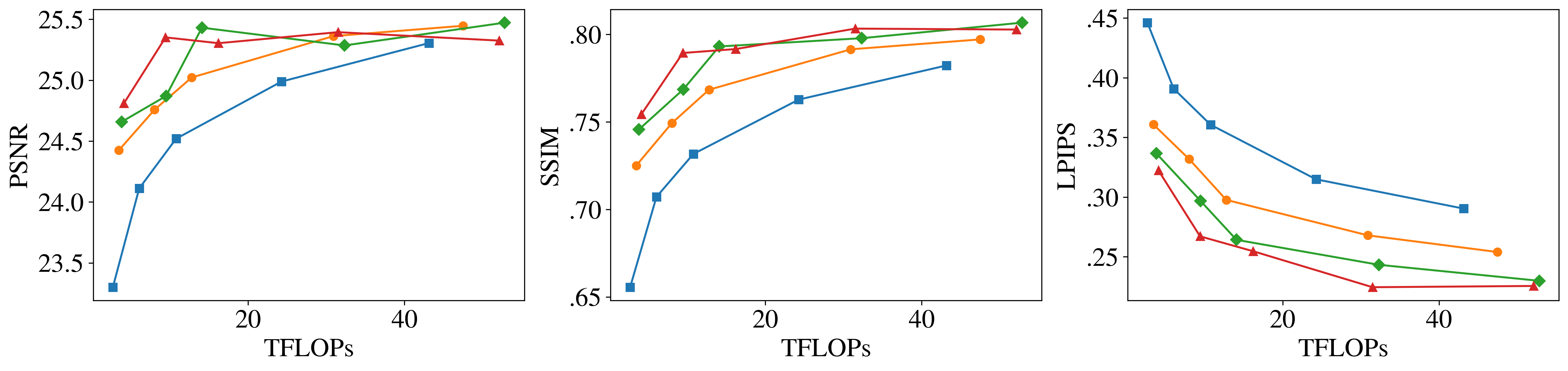}
\resizebox*{\linewidth}{2.8cm}{
\begin{tabular}{c|ccc|ccc|ccc|ccc}
\toprule
\multicolumn{1}{l|}{Units} & \multicolumn{3}{c|}{\color{blue} 1 Head ($\square$)} & \multicolumn{3}{c|}{\color{orange} 4 Heads ($\circ$)} & \multicolumn{3}{c}{\color{darkgreen} 8 Heads ($\diamond$)} & \multicolumn{3}{c}{\color{red} 16 Heads ($\triangle$)} \\
\multicolumn{1}{l|}{}      & PSNR $\uparrow$ & SSIM$\uparrow$ & LPIPS $\downarrow$ & PSNR$\uparrow$ & SSIM$\uparrow$ & LPIPS $\downarrow$ & PSNR$\uparrow$ & SSIM$\uparrow$ & LPIPS $\downarrow$ & PSNR$\uparrow$ & SSIM$\uparrow$ & LPIPS $\downarrow$ \\
\midrule
64  & 23.30 & 0.66 & 0.45 & 24.43 & 0.73 & 0.36 & 24.66 & 0.75 & 0.34 & 24.81 & 0.75 & 0.32 \\
96  & 24.11 & 0.71 & 0.39 & 24.76 & 0.75 & 0.33 & 24.87 & 0.77 & 0.30 & 25.35 & 0.79 & 0.27 \\
128  & 24.52 & 0.73 & 0.36 & 25.02 & 0.77 & 0.30 & 25.43 & 0.79 & 0.26 & 25.30 & 0.79 & 0.25 \\
192  & 24.99 & 0.76 & 0.31 & 25.36 & 0.79 & 0.27 & 25.29 & 0.80 & 0.24 & 25.39 & 0.80 & \bf{0.22} \\
256  & 25.31 & 0.78 & 0.29 & 25.45 & 0.80 & 0.25 & \bf{25.47} & \bf{0.81} & 0.23 & 25.33 & 0.80 & 0.23 \\
\bottomrule
\end{tabular}
} 
\end{center}
\vspace{-1.0em}
\caption{
Quantitative results for the ``fern'' scene.
}
\label{fig:fern}
\end{figure*}

\begin{figure*}
\footnotesize
\vspace{-1.0em}
\begin{center}
\includegraphics[width=\linewidth]{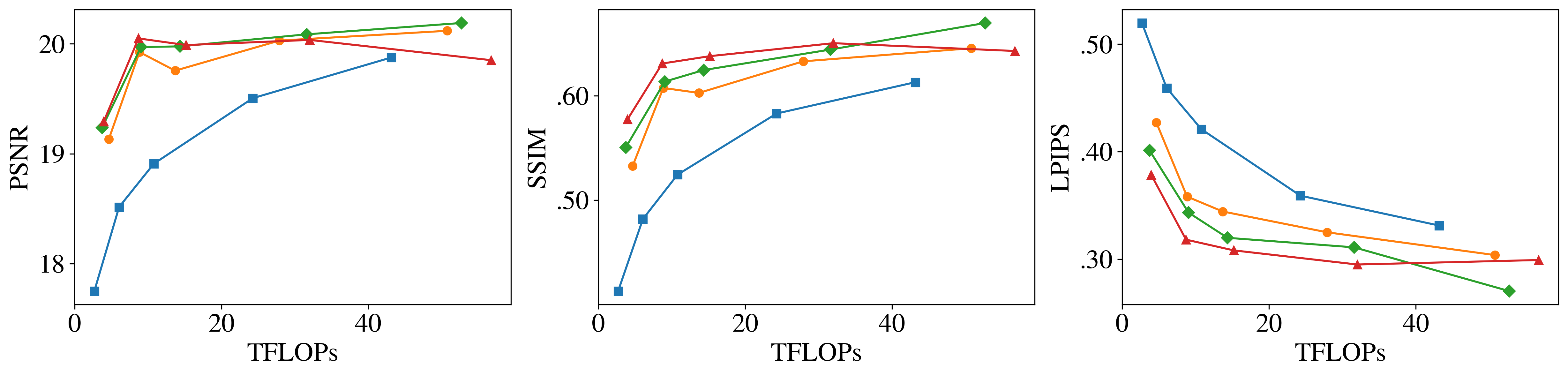}
\resizebox*{\linewidth}{2.8cm}{
\begin{tabular}{c|ccc|ccc|ccc|ccc}
\toprule
\multicolumn{1}{l|}{Units} & \multicolumn{3}{c|}{\color{blue} 1 Head ($\square$)} & \multicolumn{3}{c|}{\color{orange} 4 Heads ($\circ$)} & \multicolumn{3}{c}{\color{darkgreen} 8 Heads ($\diamond$)} & \multicolumn{3}{c}{\color{red} 16 Heads ($\triangle$)} \\
\multicolumn{1}{l|}{}      & PSNR $\uparrow$ & SSIM$\uparrow$ & LPIPS $\downarrow$ & PSNR$\uparrow$ & SSIM$\uparrow$ & LPIPS $\downarrow$ & PSNR$\uparrow$ & SSIM$\uparrow$ & LPIPS $\downarrow$ & PSNR$\uparrow$ & SSIM$\uparrow$ & LPIPS $\downarrow$ \\
\midrule
64  & 17.75 & 0.41 & 0.52 & 19.13 & 0.53 & 0.43 & 19.24 & 0.55 & 0.40 & 19.29 & 0.58 & 0.38 \\
96  & 18.51 & 0.48 & 0.46 & 19.92 & 0.61 & 0.36 & 19.97 & 0.61 & 0.34 & 20.05 & 0.63 & 0.32 \\
128  & 18.91 & 0.52 & 0.42 & 19.76 & 0.60 & 0.34 & 19.98 & 0.62 & 0.32 & 19.99 & 0.64 & 0.31 \\
192  & 19.51 & 0.58 & 0.36 & 20.03 & 0.63 & 0.33 & 20.09 & 0.64 & 0.31 & 20.04 & 0.65 & 0.30 \\
256  & 19.88 & 0.61 & 0.33 & 20.12 & 0.65 & 0.30 & \bf{20.19} & \bf{0.67} & \bf{0.27} & 19.85 & 0.64 & 0.30 \\
\bottomrule
\end{tabular}
} 
\end{center}
\vspace{-1.0em}
\caption{
Quantitative results for the ``orchids'' scene.
}
\label{fig:orchid}
\end{figure*}

\begin{figure*}
\footnotesize
\vspace{-1.0em}
\begin{center}
\includegraphics[width=\linewidth]{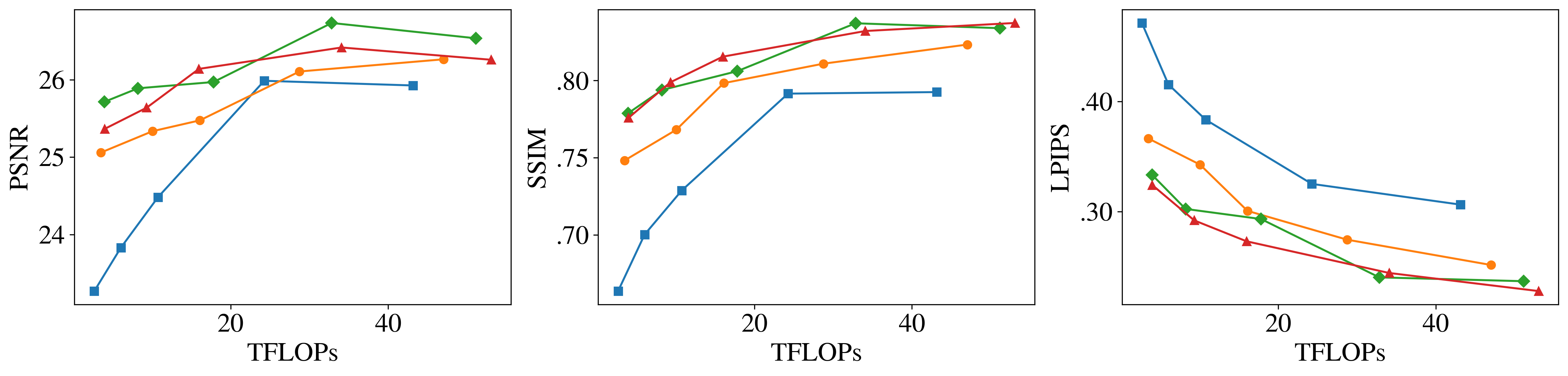}
\resizebox*{\linewidth}{2.8cm}{
\begin{tabular}{c|ccc|ccc|ccc|ccc}
\toprule
\multicolumn{1}{l|}{Units} & \multicolumn{3}{c|}{\color{blue} 1 Head ($\square$)} & \multicolumn{3}{c|}{\color{orange} 4 Heads ($\circ$)} & \multicolumn{3}{c}{\color{darkgreen} 8 Heads ($\diamond$)} & \multicolumn{3}{c}{\color{red} 16 Heads ($\triangle$)} \\
\multicolumn{1}{l|}{}      & PSNR $\uparrow$ & SSIM$\uparrow$ & LPIPS $\downarrow$ & PSNR$\uparrow$ & SSIM$\uparrow$ & LPIPS $\downarrow$ & PSNR$\uparrow$ & SSIM$\uparrow$ & LPIPS $\downarrow$ & PSNR$\uparrow$ & SSIM$\uparrow$ & LPIPS $\downarrow$ \\
\midrule
64  & 23.27 & 0.66 & 0.47 & 25.06 & 0.75 & 0.37 & 25.71 & 0.78 & 0.33 & 25.36 & 0.78 & 0.32 \\
96  & 23.83 & 0.70 & 0.42 & 25.33 & 0.77 & 0.34 & 25.89 & 0.79 & 0.30 & 25.64 & 0.80 & 0.29 \\
128  & 24.48 & 0.73 & 0.38 & 25.48 & 0.80 & 0.30 & 25.97 & 0.81 & 0.29 & 26.14 & 0.82 & 0.27 \\
192  & 25.99 & 0.79 & 0.33 & 26.11 & 0.81 & 0.27 & \bf{26.73} & \bf{0.84} & 0.24 & 26.42 & 0.83 & 0.24 \\
256  & 25.93 & 0.79 & 0.31 & 26.26 & 0.82 & 0.25 & 26.54 & 0.83 & 0.24 & 26.26 & \bf{0.84} & \bf{0.23} \\
\bottomrule
\end{tabular}
} 
\end{center}
\vspace{-1.0em}
\caption{
Quantitative results for the ``horns'' scene.
}
\label{fig:horns}
\end{figure*}

\begin{figure*}
\footnotesize
\vspace{-1.0em}
\begin{center}
\includegraphics[width=\linewidth]{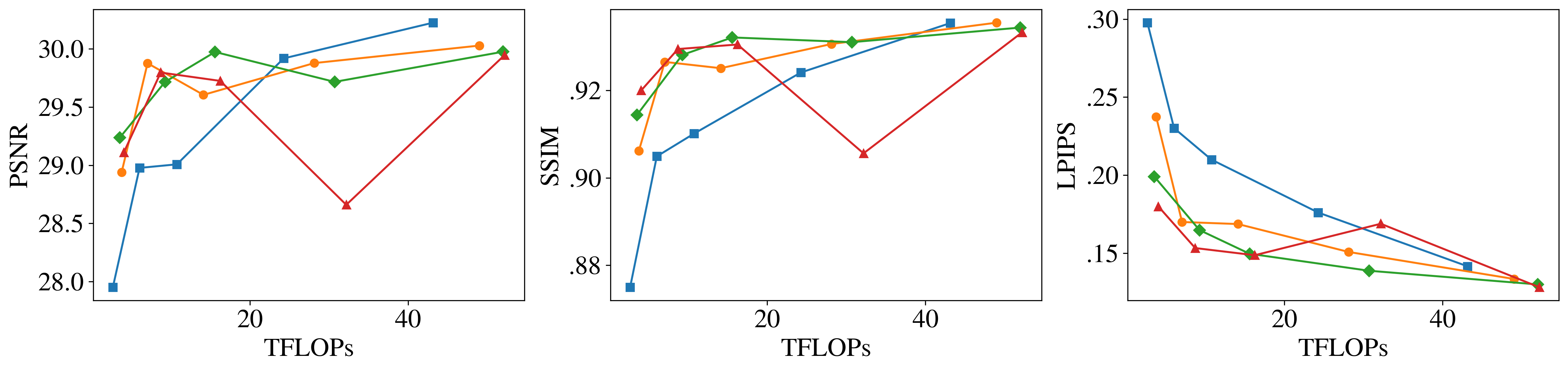}
\resizebox*{\linewidth}{2.8cm}{
\begin{tabular}{c|ccc|ccc|ccc|ccc}
\toprule
\multicolumn{1}{l|}{Units} & \multicolumn{3}{c|}{\color{blue} 1 Head ($\square$)} & \multicolumn{3}{c|}{\color{orange} 4 Heads ($\circ$)} & \multicolumn{3}{c}{\color{darkgreen} 8 Heads ($\diamond$)} & \multicolumn{3}{c}{\color{red} 16 Heads ($\triangle$)} \\
\multicolumn{1}{l|}{}      & PSNR $\uparrow$ & SSIM$\uparrow$ & LPIPS $\downarrow$ & PSNR$\uparrow$ & SSIM$\uparrow$ & LPIPS $\downarrow$ & PSNR$\uparrow$ & SSIM$\uparrow$ & LPIPS $\downarrow$ & PSNR$\uparrow$ & SSIM$\uparrow$ & LPIPS $\downarrow$ \\
\midrule
64  & 27.95 & 0.87 & 0.30 & 28.94 & 0.91 & 0.24 & 29.24 & 0.91 & 0.20 & 29.11 & 0.92 & 0.18 \\
96  & 28.98 & 0.90 & 0.23 & 29.88 & 0.93 & 0.17 & 29.72 & 0.93 & 0.16 & 29.80 & 0.93 & 0.15 \\
128  & 29.01 & 0.91 & 0.21 & 29.60 & 0.93 & 0.17 & 29.97 & 0.93 & 0.15 & 29.72 & 0.93 & 0.15 \\
192  & 29.92 & 0.92 & 0.18 & 29.88 & 0.93 & 0.15 & 29.72 & 0.93 & 0.14 & 28.66 & 0.91 & 0.17 \\
256  & \bf{30.22} & \bf{0.94} & 0.14 & 30.03 & \bf{0.94} & \bf{0.13} & 29.97 & 0.93 & \bf{0.13} & 29.95 & 0.93 & \bf{0.13} \\
\bottomrule
\end{tabular}
} 
\end{center}
\vspace{-1.0em}
\caption{
Quantitative results for the ``room'' scene.
}
\label{fig:room}
\end{figure*}

\begin{figure*}
\footnotesize
\vspace{-1.0em}
\begin{center}
\includegraphics[width=\linewidth]{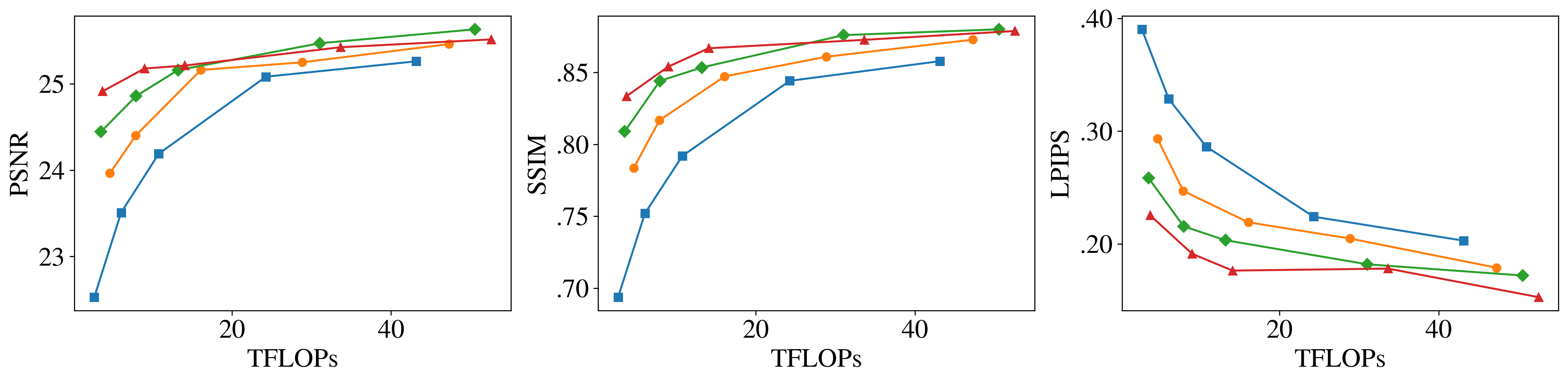}
\resizebox*{\linewidth}{2.8cm}{
\begin{tabular}{c|ccc|ccc|ccc|ccc}
\toprule
\multicolumn{1}{l|}{Units} & \multicolumn{3}{c|}{\color{blue} 1 Head ($\square$)} & \multicolumn{3}{c|}{\color{orange} 4 Heads ($\circ$)} & \multicolumn{3}{c}{\color{darkgreen} 8 Heads ($\diamond$)} & \multicolumn{3}{c}{\color{red} 16 Heads ($\triangle$)} \\
\multicolumn{1}{l|}{}      & PSNR $\uparrow$ & SSIM$\uparrow$ & LPIPS $\downarrow$ & PSNR$\uparrow$ & SSIM$\uparrow$ & LPIPS $\downarrow$ & PSNR$\uparrow$ & SSIM$\uparrow$ & LPIPS $\downarrow$ & PSNR$\uparrow$ & SSIM$\uparrow$ & LPIPS $\downarrow$ \\
\midrule
64  & 22.53 & 0.69 & 0.39 & 23.96 & 0.78 & 0.29 & 24.44 & 0.81 & 0.26 & 24.91 & 0.83 & 0.23 \\
96  & 23.51 & 0.75 & 0.33 & 24.40 & 0.82 & 0.25 & 24.86 & 0.84 & 0.22 & 25.18 & 0.85 & 0.19 \\
128  & 24.19 & 0.79 & 0.29 & 25.16 & 0.85 & 0.22 & 25.16 & 0.85 & 0.20 & 25.21 & 0.87 & 0.18 \\
192  & 25.08 & 0.84 & 0.22 & 25.25 & 0.86 & 0.20 & 25.47 & \bf{0.88} & 0.18 & 25.42 & 0.87 & 0.18 \\
256  & 25.26 & 0.86 & 0.20 & 25.46 & 0.87 & 0.18 & \bf{25.63} & \bf{0.88} & 0.17 & 25.51 & \bf{0.88} & \bf{0.15} \\
\bottomrule
\end{tabular}
} 
\end{center}
\vspace{-1.0em}
\caption{
Quantitative results for the ``trex'' scene.
}
\label{fig:trex}
\end{figure*}

\begin{figure*}
\footnotesize
\vspace{-1.0em}
\begin{center}
\includegraphics[width=\linewidth]{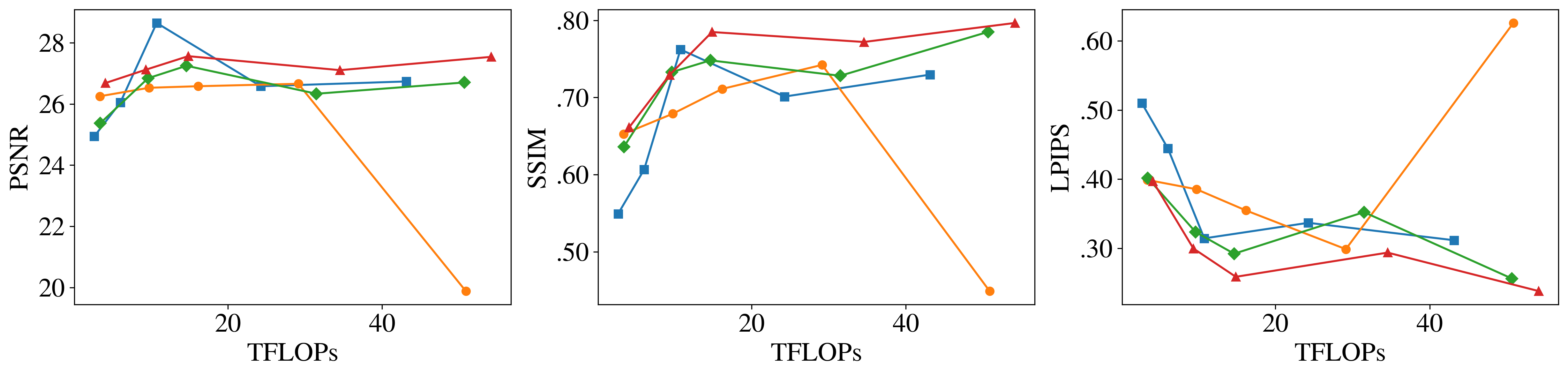}
\resizebox*{\linewidth}{2.8cm}{
\begin{tabular}{c|ccc|ccc|ccc|ccc}
\toprule
\multicolumn{1}{l|}{Units} & \multicolumn{3}{c|}{\color{blue} 1 Head ($\square$)} & \multicolumn{3}{c|}{\color{orange} 4 Heads ($\circ$)} & \multicolumn{3}{c}{\color{darkgreen} 8 Heads ($\diamond$)} & \multicolumn{3}{c}{\color{red} 16 Heads ($\triangle$)} \\
\multicolumn{1}{l|}{}      & PSNR $\uparrow$ & SSIM$\uparrow$ & LPIPS $\downarrow$ & PSNR$\uparrow$ & SSIM$\uparrow$ & LPIPS $\downarrow$ & PSNR$\uparrow$ & SSIM$\uparrow$ & LPIPS $\downarrow$ & PSNR$\uparrow$ & SSIM$\uparrow$ & LPIPS $\downarrow$ \\
\midrule
64  & 24.94 & 0.55 & 0.51 & 26.25 & 0.65 & 0.40 & 25.37 & 0.64 & 0.40 & 26.68 & 0.66 & 0.40 \\
96  & 26.05 & 0.61 & 0.44 & 26.53 & 0.68 & 0.39 & 26.84 & 0.73 & 0.32 & 27.12 & 0.73 & 0.30 \\
128  & \bf{28.65} & 0.76 & 0.31 & 26.58 & 0.71 & 0.35 & 27.25 & 0.75 & 0.29 & 27.56 & 0.78 & 0.26 \\
192  & 26.58 & 0.70 & 0.34 & 26.66 & 0.74 & 0.30 & 26.33 & 0.73 & 0.35 & 27.10 & 0.77 & 0.29 \\
256  & 26.74 & 0.73 & 0.31 & 19.88 & 0.45 & 0.63 & 26.70 & 0.79 & 0.26 & 27.54 & \bf{0.80} & \bf{0.24} \\
\bottomrule
\end{tabular}
} 
\end{center}
\vspace{-1.0em}
\caption{
Quantitative results for the ``fortress'' scene.
}
\label{fig:fortress}
\end{figure*}

\begin{figure*}
\footnotesize
\vspace{-1.0em}
\begin{center}
\includegraphics[width=\linewidth]{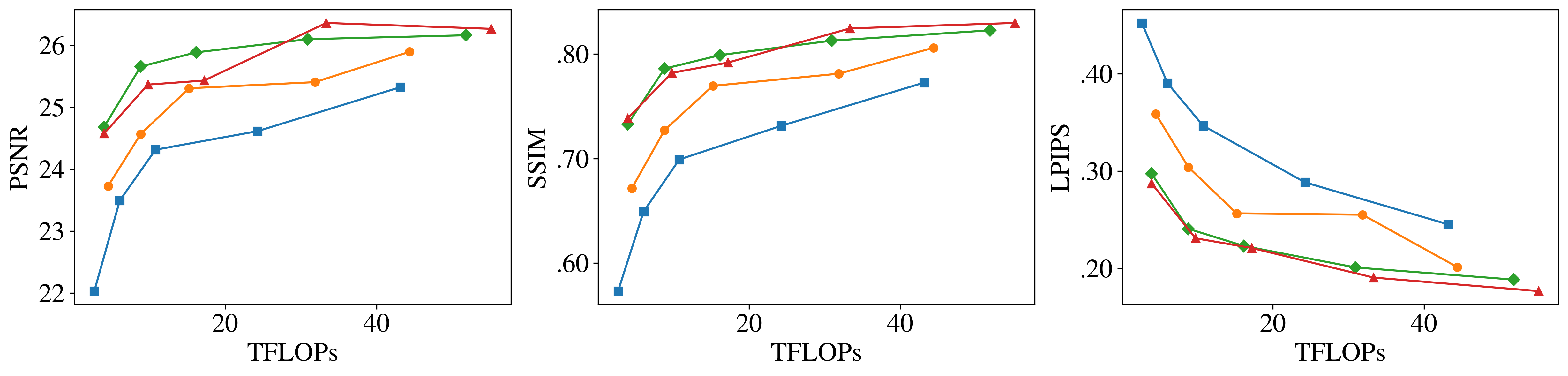}
\resizebox*{\linewidth}{2.8cm}{
\begin{tabular}{c|ccc|ccc|ccc|ccc}
\toprule
\multicolumn{1}{l|}{Units} & \multicolumn{3}{c|}{\color{blue} 1 Head ($\square$)} & \multicolumn{3}{c|}{\color{orange} 4 Heads ($\circ$)} & \multicolumn{3}{c}{\color{darkgreen} 8 Heads ($\diamond$)} & \multicolumn{3}{c}{\color{red} 16 Heads ($\triangle$)} \\
\multicolumn{1}{l|}{}      & PSNR $\uparrow$ & SSIM$\uparrow$ & LPIPS $\downarrow$ & PSNR$\uparrow$ & SSIM$\uparrow$ & LPIPS $\downarrow$ & PSNR$\uparrow$ & SSIM$\uparrow$ & LPIPS $\downarrow$ & PSNR$\uparrow$ & SSIM$\uparrow$ & LPIPS $\downarrow$ \\
\midrule
64  & 22.03 & 0.57 & 0.45 & 23.73 & 0.67 & 0.36 & 24.68 & 0.73 & 0.30 & 24.57 & 0.74 & 0.29 \\
96  & 23.49 & 0.65 & 0.39 & 24.57 & 0.73 & 0.30 & 25.66 & 0.79 & 0.24 & 25.36 & 0.78 & 0.23 \\
128  & 24.31 & 0.70 & 0.35 & 25.30 & 0.77 & 0.26 & 25.88 & 0.80 & 0.22 & 25.43 & 0.79 & 0.22 \\
192  & 24.61 & 0.73 & 0.29 & 25.40 & 0.78 & 0.26 & 26.10 & 0.81 & 0.20 & \bf{26.36} & 0.82 & 0.19 \\
256  & 25.32 & 0.77 & 0.25 & 25.89 & 0.81 & 0.20 & 26.16 & 0.82 & 0.19 & 26.26 & \bf{0.83} & \bf{0.18} \\
\bottomrule
\end{tabular}
} 
\end{center}
\vspace{-1.0em}
\caption{
Quantitative results for the ``flower'' scene.
}
\label{fig:flower}
\end{figure*}

\begin{figure*}[t]
\begin{center}
\includegraphics[width=\linewidth]{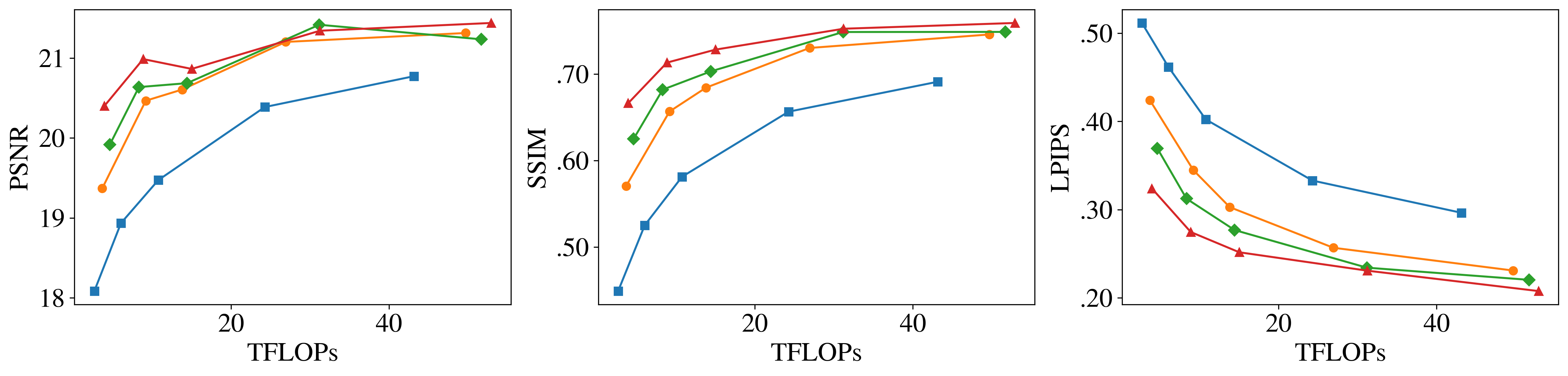}
\resizebox*{\linewidth}{2.8cm}{
\begin{tabular}{c|ccc|ccc|ccc|ccc}
\toprule
\multicolumn{1}{l|}{Units} & \multicolumn{3}{c|}{\color{blue} 1 Head ($\square$)} & \multicolumn{3}{c|}{\color{orange} 4 Heads ($\circ$)} & \multicolumn{3}{c}{\color{darkgreen} 8 Heads ($\diamond$)} & \multicolumn{3}{c}{\color{red} 16 Heads ($\triangle$)} \\
\multicolumn{1}{l|}{}      & PSNR $\uparrow$ & SSIM$\uparrow$ & LPIPS $\downarrow$ & PSNR$\uparrow$ & SSIM$\uparrow$ & LPIPS $\downarrow$ & PSNR$\uparrow$ & SSIM$\uparrow$ & LPIPS $\downarrow$ & PSNR$\uparrow$ & SSIM$\uparrow$ & LPIPS $\downarrow$ \\
\midrule
64  & 18.08 & 0.45 & 0.51 & 19.37 & 0.57 & 0.42 & 19.92 & 0.63 & 0.37 & 20.40 & 0.67 & 0.32 \\
96  & 18.93 & 0.53 & 0.46 & 20.47 & 0.66 & 0.34 & 20.64 & 0.68 & 0.31 & 20.99 & 0.71 & 0.27 \\
128  & 19.47 & 0.58 & 0.40 & 20.61 & 0.68 & 0.30 & 20.69 & 0.70 & 0.28 & 20.86 & 0.73 & 0.25 \\
192  & 20.39 & 0.66 & 0.33 & 21.20 & 0.73 & 0.26 & 21.42 & 0.75 & 0.23 & 21.34 & 0.75 & 0.23 \\
256  & 20.77 & 0.69 & 0.30 & 21.31 & 0.75 & 0.23 & 21.23 & 0.75 & 0.22 & \bf{21.44} & \bf{0.76} & \bf{0.21} \\
\bottomrule
\end{tabular}
} 
\end{center}
\caption{
Quantitative results for the ``leaves'' scene.
}
\label{fig:leaves}
\end{figure*}

\paragraph{Per-scene results -- Figures~\ref{fig:fern}--\ref{fig:leaves}}
In this section we provide further comprehensive experiments on the real capture scenes showing how DeRFs with up to 16 heads perform in terms of reconstruction quality and inference cost.

We find that DeRF models provide the best quality-computation trade-off, for all cases in terms of LPIPS, and almost all cases for PSNR and SSIM.
Moreover, for the strong majority of scenes, highly decomposed DeRF models give the best results, especially for the perceptual and structural metrics.
We also find the advantage of DeRF is most consistent for scenes where the absolute error is higher, i.e. where gains in performance are needed the most.
All experiments in this section use the same sample counts as \cite{nerf}.

\paragraph{Variance Test -- Table~\ref{tab:variance}}
Additionally, to help explain the increasingly chaotic nature of the graphs as quality increases, we perform an experiment in which we train eight identical DeRF models on the scene with the highest reconstruction accuracy (the ``room'' scene).
In this test we find a significant difference in results depending only on initialization, especially for PSNR.
This suggests that, while the current experiments already provide enough evidence for the efficacy of DeRF, extracting robust statistics via multiple runs may deliver a more conclusive answer at the expense of compute.
We note however, our results already demonstrate that decomposing the scene is almost always better, especially in terms of perceptual metrics, and when less computation is used for inference.

\paragraph{Video supplementary}
We further direct the interested readers to the video supplementary that demonstrates the rendering quality and the decomposition in 3D.

\begin{table}[b]
\vspace{-1.0em}
\footnotesize
\setlength{\tabcolsep}{12pt}
\begin{center}
\begin{tabular}{l|ccc}
\toprule
 & PSNR$\uparrow$ & SSIM$\uparrow$ & LPIPS$\downarrow$ \\ 
 \midrule
\MethodName & \bf{28.55} & \bf{0.89} & \bf{0.25} \\
\MethodName-Grid & 28.07 & 0.88 & 0.28 \\
\bottomrule
\end{tabular}
\end{center}
\caption{
{\bf Voronoi vs grid decomposition -- }
Quantitative results for two equal-capacity 64-head, 32-unit models trained on the ``room'' scene.
The DeRF model uses our Voronoi decomposition and the DeRF-Grid uses an untrained decomposition where the cells are arranged in a 4$\times$4$\times$4 grid.
Our Voronoi decomposition brings significant benefit in terms of rendering quality.
}
\label{tab:gridablation}
\end{table}
\begin{table}[b]
\footnotesize
\setlength{\tabcolsep}{12pt}
\begin{center}
\begin{tabular}{l|ccc}
\toprule
 & PSNR$\uparrow$ & SSIM$\uparrow$ & LPIPS$\downarrow$ \\ 
 \midrule
Minimum & 29.15 & 0.92 & \bf{0.15} \\
Maximum & \bf{29.86} & \bf{0.93} & 0.16 \\
\bottomrule
\end{tabular}
\end{center}
\caption{
{\bf Variation across initialization --}
We report the variation in rendering quality resulting from different initializations by showing the minimum and maximum metrics, achieved by a set of 16-head, 128-unit DeRF models trained on the ``room'' scene.
}
\label{tab:variance}
\end{table}


\clearpage

\begin{figure}
\footnotesize
\begin{center}
\includegraphics[width=\linewidth]{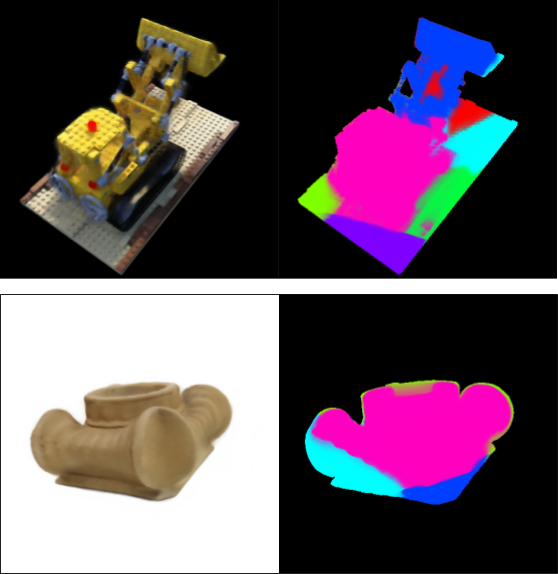}
\end{center}
\caption{
\textbf{Synthetic Data} --
Renders and decompositions for the NeRF~\cite{nerf} ``lego'' scene (top) and DeepVoxels~\cite{deepvoxels} ``greek'' scene (bottom).
}
\label{fig:synthetic}
\end{figure}

\clearpage
\section{Synthetic data -- \Figure{synthetic}}

While we are mainly interested in photorealistic scenes, our method can also learn decomposed models from synthetic data.
To show this we provide renders and visualize the decompositions for DeRF models trained on two synthetic scenes.
Each model uses 192 units and 8 heads.

\end{document}